\documentclass{article} 
\usepackage{iclr2021_conference,times}

\usepackage{amsmath,amsfonts,bm}

\def\eqref#1{equation~\ref{#1}}

\def\1{\bm{1}}

\DeclareMathAlphabet{\mathsfit}{\encodingdefault}{\sfdefault}{m}{sl}
\SetMathAlphabet{\mathsfit}{bold}{\encodingdefault}{\sfdefault}{bx}{n}

\usepackage{hyperref}
\usepackage{url}
\usepackage{times}
\usepackage{epsfig}
\usepackage{graphicx}
\usepackage{amsmath}
\usepackage{amssymb}
\usepackage{bm}

\usepackage{subcaption}
\usepackage{multirow}
\usepackage{multicol}
\usepackage{ctable}
\usepackage{float}
\usepackage{pifont}
\usepackage{wrapfig}
\usepackage{upgreek}
\usepackage{soul}

\title{Revisiting Dynamic Convolution via Matrix Decomposition}

\newcommand{\tabincell}[2]{\begin{tabular}
{@{}#1@{}}#2\end{tabular}}

\newcommand{\xmark}{\ding{86}}

\author{Yunsheng Li\textsuperscript{1}, Yinpeng Chen\textsuperscript{2}, Xiyang Dai\textsuperscript{2}, Mengchen Liu\textsuperscript{2}, Dongdong Chen\textsuperscript{2}, Ye Yu\textsuperscript{2}, \\
\textbf{Lu Yuan\textsuperscript{2}, Zicheng Liu\textsuperscript{2}, Mei Chen\textsuperscript{2}, Nuno Vasconcelos\textsuperscript{1}} \\
\textsuperscript{1}  Department of Electrical and Computer Engineering, University of California San Diego \\
\textsuperscript{2} Microsoft \\
\texttt{yul554@ucsd.edu, \{yiche,xidai,mengcliu,dochen\}@microsoft.com} \\
\texttt{\{Yu.Ye,luyuan,zliu,Mei.Chen\}@microsoft.com, nvasconcelos@ucsd.edu}
}
\iclrfinalcopy 
\begin{document}

\maketitle

\begin{abstract}
Recent research in dynamic convolution shows substantial performance boost for efficient CNNs, due to the adaptive aggregation of $K$ static convolution kernels. It has two limitations: (a) it increases the number of convolutional weights by $K$-times, and (b) the joint optimization of dynamic attention and static convolution kernels is challenging. In this paper, we revisit it from a new perspective of matrix decomposition and reveal the key issue is that dynamic convolution applies dynamic attention over channel groups after projecting into a higher dimensional latent space. To address this issue, we propose dynamic channel fusion to replace dynamic attention over channel groups. Dynamic channel fusion not only enables significant dimension reduction of the latent space, but also mitigates the joint optimization difficulty. As a result, our method is easier to train and requires significantly fewer parameters without sacrificing accuracy. Source code is at \href{https://github.com/liyunsheng13/dcd}{https://github.com/liyunsheng13/dcd}.

\end{abstract}
\section{Introduction}

Dynamic convolution \citep{Yang2019CondConvCP, Chen2019DynamicCA} has recently become popular for the implementation of  light-weight networks \citep{howard2017mobilenets, Zhang_2018_CVPR}. Its ability to achieve significant performance gains with negligible computational cost has motivated its adoption for multiple vision tasks \citep{Su_2020_eccv_DynamicGC,Chen2020DynamicRC, Ma_2020_eccv_WeightNetRT, Tian_2020_eccv_ConditionalCF}.
The basic idea is to aggregate multiple convolution kernels dynamically, according to an input dependent attention mechanism, into a convolution weight matrix
\begin{align}
\bm{W}(\bm{x})=\sum_{k=1}^K\pi_k(\bm{x})\bm{W}_k \;\;\;
\text{s.t.} \;\;\; 0 \leq \pi_k(\bm{x}) \leq 1, 
\sum_{k=1}^K \pi_k(\bm{x}) = 1,
\label{eq:dynamic-conv}
\end{align}
where $K$ convolution kernels $\{\bm{W}_k\}$ are aggregated linearly with attention scores $\{\pi_k(\bm{x})\}$. 

Dynamic convolution has two main limitations: (a) lack of compactness, due to the use of $K$ kernels, and (b) a challenging joint optimization of attention scores $\{\pi_k(\bm{x})\}$ and static kernels $\{\bm{W}_k\}$. 
\cite{Yang2019CondConvCP} proposed the use of a sigmoid layer to generate attention scores $\{\pi_k(\bm{x})\}$, leading to a significantly large space for the convolution kernel $\bm{W}(\bm{x})$ that makes the learning of attention scores $\{\pi_k(\bm{x})\}$ difficult. \cite{Chen2019DynamicCA} replaced the sigmoid layer with a softmax function to compress the kernel space. However, small attention scores $\pi_k$ output by the softmax make the corresponding kernels $\bm{W}_k$ difficult to learn, especially in early training epochs, slowing training convergence. To mitigate these limitations, these two methods require additional constraints. For instance, 
\cite{Chen2019DynamicCA} uses a large temperature in the softmax function to encourage near-uniform attention. 

\begin{figure}[t]
\centering
\includegraphics[width=.95\linewidth]{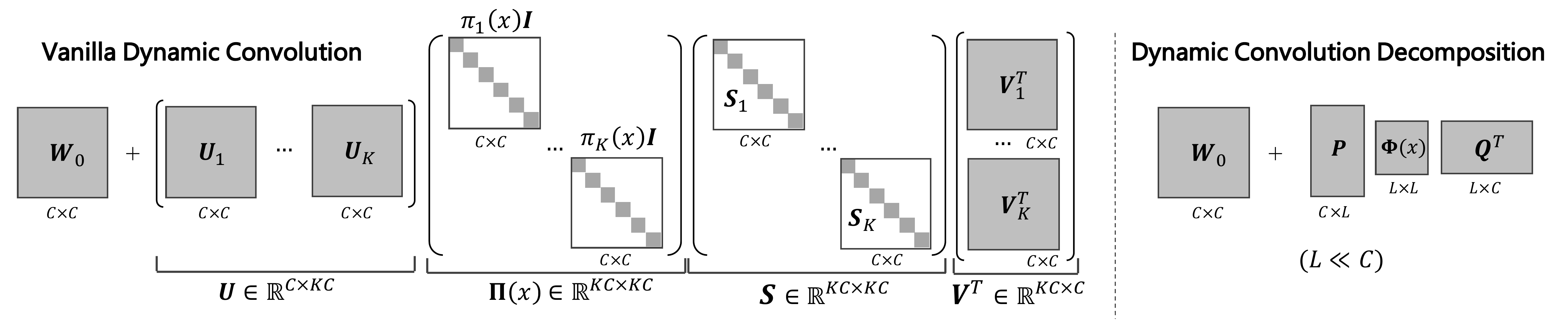}
\caption{Dynamic convolution via matrix decomposition. 
\textbf{Left}: Reformulating the vanilla dynamic convolution by matrix decomposition (see Eq. \ref{eq:dynamic-decom}). It applies \textbf{\textit{dynamic attention $\bm{\Pi}(\bm{x})$ over channel groups}} in a \textbf{\textit{high dimensional space}} ($\bm{S}\bm{V}^T\bm{x} \in \mathbb{R}^{KC}$). \textbf{Right}: proposed dynamic convolution decomposition, which applies \textbf{\textit{dynamic channel fusion}} $\bm{\Phi}(\bm{x})$ in a \textbf{\textit{low dimensional space}} ($\bm{Q}^T\bm{x} \in \mathbb{R}^L$, $L\ll C$), resulting in a more compact model. 
}
\label{fig:teaser_dyconv}
\vspace{-1.0em}
\end{figure}

In this work, we revisit the two limitations via matrix decomposition. To expose the limitations, we reformulate dynamic convolution in terms of a set of residuals, re-defining the static kernels as 
\begin{equation}
    \bm{W}_k = \bm{W}_0 + \Delta\bm{W}_k, \quad k \in \{1, \ldots, K\}
\end{equation}
where $\bm{W}_0 = \frac{1}{K}\sum_{k=1}^K \bm{W}_k$ is the average kernel and $\Delta\bm{W}_k = \bm{W}_k - \bm{W}_0$ a residual weight matrix. Further decomposing the latter with an SVD, $\Delta \bm{W}_k=\bm{U}_k\bm{S}_k\bm{V}_k^T$, leads to
\begin{align}
\bm{W}(\bm{x})=\sum_{k=1}^K \pi_k(\bm{x})\bm{W}_0 + \sum_{k=1}^K\pi_k(\bm{x})\bm{U}_k\bm{S}_k\bm{V}_k^T 
=\bm{W}_0 + \bm{U}\bm{\Pi}(\bm{x})\bm{S}\bm{V}^T, 
\label{eq:dynamic-decom}
\end{align}

where 
$\bm{U}=[\bm{U}_1,\dots,\bm{U}_K]$, $\bm{S}=diag(\bm{S}_1,\dots,\bm{S}_K)$, $\bm{V}=[\bm{V}_1,\dots,\bm{V}_K]$, and $\bm{\Pi}(\bm{x})$ stacks attention scores diagonally as $\bm{\Pi}(\bm{x})=diag(\pi_1(\bm{x})\bm{I},\dots,\pi_K(\bm{x})\bm{I})$, where ${\bm{I}}$ is an identity matrix. This decomposition, illustrated in Figure \ref{fig:teaser_dyconv},
shows that the dynamic behavior of $\bm{W}(\bm{x})$ is implemented by the dynamic residual $\bm{U}\bm{\Pi}(\bm{x})\bm{S}\bm{V}^T$, which projects the input $\bm{x}$ to a higher dimensional space $\bm{S}\bm{V}^T\bm{x}$ (from $C$ to $KC$ channels), applies dynamic attention $\bm{\Pi}(\bm{x})$ over channel groups, and reduces the dimension back to $C$ channels, through multiplication by $\bm{U}$. 
This suggests that the limitations of vanilla dynamic convolution are due to the use of \textit{attention over channel groups,} which induces a high dimensional latent space, leading to small attention values that may suppress the learning of the corresponding channels.

To address this issue, we propose a \textit{dynamic convolution decomposition} (DCD), that replaces dynamic attention over channel groups with \textit{dynamic channel fusion}. The latter is based on a full dynamic matrix $\bm{\Phi}(\bm{x})$, of which each element $\phi_{i,j}(\bm{x})$ is a function of input $\bm{x}$. As shown in Figure \ref{fig:teaser_dyconv}-(right), the dynamic residual is implemented as the product $\bm{P}\bm{\Phi}(\bm{x})\bm{Q}^T$ of $\bm{\Phi}(\bm{x})$ and two static matrices $\bm{P},\bm{Q}$, such that $\bm{Q}$ compresses the input into a low dimensional latent space, $\bm{\Phi}(\bm{x})$ dynamically fuses the channels in this space, and $\bm{P}$ expands the number of channels to the output space. The key innovation is that dynamic channel fusion with $\bm{\Phi}(\bm{x})$ enables a significant dimensionality reduction of the latent space ($\bm{Q}^T\bm{x} \in \mathbb{R}^{L}$, $L \ll C$). Hence the number of parameters in $\bm{P}, \bm{Q}$ is significantly reduced, when compared to $\bm{U}, \bm{V}$ of Eq. \ref{eq:dynamic-decom}, resulting in a more compact model. 
Dynamic channel fusion also mitigates the joint optimization challenge of vanilla dynamic convolution, as each column of $\bm{P}, \bm{Q}$ is associated with multiple dynamic coefficients of $\bm{\Phi}(\bm{x})$. Hence, a few dynamic coefficients of small value are not sufficient to suppress the learning of static matrices $\bm{P}, \bm{Q}$. 
Experimental results show that DCD both significantly reduces the number of parameters 
and achieves higher accuracy than vanilla dynamic convolution, without requiring the additional constraints of \citep{Yang2019CondConvCP, Chen2019DynamicCA}.

\section{Related Work}
\textbf{Efficient CNNs:} MobileNet \citep{howard2017mobilenets, sandler2018mobilenetv2, howard2019mbnetv3} decomposes $k \times k$ convolution into a depthwise and a pointwise convolution. ShuffleNet \citep{Zhang_2018_CVPR, ma_2018_ECCV} uses group convolution and channel shuffle to further simplify pointwise convolution. Further improvements of these architectures have been investigated recently. EfficientNet \citep{tan-ICML19-efficientnet, Tan_2020_CVPR} finds a proper relationship between input resolution and width/depth of the network. \cite{Tan-bmvc2019-mixconv} mix up multiple kernel sizes in a single convolution. \cite{Chen_2020_CVPR_addernet} trades massive multiplications for much cheaper additions. \cite{Han_2020_CVPR_ghostnet} applies a series of cheap linear transformations to generate ghost feature maps. \cite{Daquan_2020_ECCV_RethinkingBS} flips the structure of inverted residual blocks to alleviate information loss.
\cite{yu2018slimmable} and \cite{Cai2019OnceFA} train one network that supports multiple sub-networks of different complexities. 

\textbf{Matrix Decomposition:} 
\cite{lebedev2014speeding} and \cite{denton2014exploiting} use Canonical Polyadic decomposition (CPD) of convolution kernels to speed up networks, while \cite{kim2015compression} investigates Tucker decompositions for the same purpose.
More recently, \cite{kossaifi2020factorized} combines tensor decompositions with MobileNet to design efficient higher-order networks for video tasks, while \cite{phan2020stable} proposes a stable CPD to deal with degeneracies of tensor decompositions during network training. Unlike DCD, which decomposes a convolutional kernel  dynamically by adapting the core matrix to the input, these works all rely on static decompositions.

\textbf{Dynamic Neural Networks:} Dynamic networks boost representation power by adapting parameters or activation functions to the input. \cite{Ha2017HyperNetworks} uses a secondary network to generate parameters for the main network. \cite{Hu_2018_CVPR} reweights channels by squeezing global context. \cite{Li_2019_CVPR_SKNet} adapts attention over kernels of different sizes. Dynamic convolution \citep{Yang2019CondConvCP,Chen2019DynamicCA} aggregates multiple convolution kernels based on attention. \cite{Ma_2020_eccv_WeightNetRT} uses grouped fully connected layer to generate convolutional weights directly.
\cite{Chen2020DynamicRC} extends dynamic convolution from spatial agnostic to spatial specific. \cite{Su_2020_eccv_DynamicGC} proposes dynamic group convolution that adaptively selects input channels to form groups. \cite{Tian_2020_eccv_ConditionalCF} applies dynamic convolution to instance segmentation. \cite{Chen2020DynamicReLU} adapts slopes and intercepts of two linear functions in ReLU \citep{NairH10Relu,JarrettKRL09Relu}. 

\section{Dynamic Convolution Decomposition}
In this section, we introduce the \textit{dynamic convolution decomposition} proposed to address the limitations of vanilla dynamic convolution. For conciseness, we assume a kernel $\bm{W}$ with the same number of input and output channels ($C_{in}=C_{out}=C$) and ignore bias terms. 
We focus on $1 \times 1$ convolution in this section and generalize the procedure to $k \times k$ convolution in the following section.

\subsection{Revisiting Vanilla Dynamic Convolution}
Vanilla dynamic convolution aggregates $K$ convolution kennels $\{\bm{W}_k\}$ with attention scores $\{\pi_k(\bm{x})\}$ (see Eq. \ref{eq:dynamic-conv}). It can be reformulated as adding a dynamic residual to a static kernel, and the dynamic residual can be further decomposed by SVD (see Eq. \ref{eq:dynamic-decom}), as shown in Figure \ref{fig:teaser_dyconv}.  
This has two limitations. First, the model is not compact. 
Essentially, \textit{it expands the number of channels by a factor of $K$ and applies dynamic attention over $K$ channel groups.}
The dynamic residual $\bm{U}\bm{\Pi}(\bm{x})\bm{S}\bm{V}^T$ is a $C \times C$ matrix, of maximum rank $C$, but sums $KC$ rank-1 matrices, since
\begin{align}
\bm{W}(\bm{x})=\bm{W}_0 + \bm{U}\bm{\Pi}(\bm{x})\bm{S}\bm{V}^T 
=\bm{W}_0 + \sum_{i=1}^{KC}\pi_{\lceil i/C \rceil}(\bm{x})\bm{u}_{i}s_{i,i}\bm{v}_{i}^T,
\label{eq:dynamic-attention-redidual-svd2}
\end{align}
where $\bm{u}_i$ is the $i^{th}$ column vector of matrix $\bm{U}$, $\bm{v}_i$ is the $i^{th}$ column vector of matrix $\bm{V}$, $s_{i, i}$ is the $i^{th}$ diagonal entry of matrix $\bm{S}$ and $\lceil \cdot \rceil$ is ceiling operator. The static basis vectors $\bm{u}_i$ and $\bm{v}_i$ are not shared across different rank-1 matrices ($\pi_{\lceil i/C \rceil}(\bm{x})\bm{u}_{i}s_{i,i}\bm{v}_{i}^T$). This results in model redundancy. Second, it is difficult to jointly optimize static matrices $\bm{U}$, $\bm{V}$ and dynamic attention $\bm{\Pi}(\bm{x})$. This is because a small attention score $\pi_{\lceil i/C \rceil}$ may suppress the learning of corresponding columns $\bm{u}_i$, $\bm{v}_i$ in $\bm{U}$ and $\bm{V}$, especially in early training epochs  (as shown in \cite{Chen2019DynamicCA}).

\subsection{Dynamic Channel Fusion}

We propose to address the limitations of the vanilla dynamic convolution with a dynamic channel fusion mechanism, implemented with a full matrix $\bm{\Phi}(\bm{x})$, where each element $\phi_{i,j}(\bm{x})$ is a function of input $\bm{x}$. $\bm{\Phi}(\bm{x})$ is a $L \times L$ matrix, dynamically fusing channels in the latent space $\mathbb{R}^L$. The key idea is to significantly reduce dimensionality in the latent space, $L \ll C$, to enable a more compact model. Dynamic convolution is implemented with dynamic channel fusion using
\begin{align}
\bm{W}(\bm{x})= \bm{W}_0 + \bm{P}\bm{\Phi}(\bm{x})\bm{Q}^T
=\bm{W}_0 + \sum_{i=1}^{L} \sum_{j=1}^{L} \bm{p}_i\phi_{i,j}(\bm{x})\bm{q}_j^T,
\label{eq:dynamic-residual}
\end{align}
where $\bm{Q} \in \mathbb{R}^{C \times L}$ compresses the input into a low dimensional space ($\bm{Q}^T\bm{x} \in \mathbb{R}^{L}$), the resulting $L$ channels are fused dynamically by $\bm{\Phi}(\bm{x}) \in \mathbb{R}^{L \times L}$ and expanded to the number of output channels by  $\bm{P} \in \mathbb{R}^{C \times L}$. This is denoted as \textbf{dynamic convolution decomposition} (DCD). The dimension $L$ of the latent space  is constrained by $L^2<C$. The default value of $L$ in this paper is empirically set to $\lfloor \frac{C}{2^{\lfloor \log_2\sqrt{C} \rfloor}} \rfloor$, which means dividing $C$ by 2 repeatedly until it is less than $\sqrt{C}$.

With this new design, the number of static parameters is significantly reduced (i.e. $LC$ parameters in $\bm{P}$ or $\bm{Q}$ \textit{v.s.} $KC^2$ parameters in $\bm{U}$ or $\bm{V}$, $L < \sqrt{C}$), resulting in a more compact model. 
Mathematically, the dynamic residual $\bm{P}\bm{\Phi}(\bm{x})\bm{Q}^T$ sums $L^2$ rank-1 matrices $\bm{p}_i\phi_{i,j}(\bm{x})\bm{q}_j^T$, where $\bm{p}_i$ is the $i^{th}$ column vector of $\bm{P}$, and $\bm{q}_j$ is the $j^{th}$ column vector of $\bm{Q}$. The constraint $L^2<C$, guarantees that this number ($L^2$) is much smaller than the counterpart ($KC$) of vanilla dynamic convolution (see Eq. \ref{eq:dynamic-attention-redidual-svd2}).
Nevertheless, due to the use of a full matrix, dynamic channel fusion $\bm{\Phi}(\bm{x})$ retains the representation power needed to achieve good classification performance.

DCD also mitigates the joint optimization difficulty. Since each column of $\bm{P}$ (or $\bm{Q}$) is associated with multiple dynamic coefficients (e.g. $\bm{p}_i$ is related to $\phi_{i, 1},\dots, \phi_{i, L}$), it is unlikely that the learning of $\bm{p}_i$ is suppressed by a few dynamic coefficients of small value.

In summary, DCD performs dynamic aggregation differently from vanilla dynamic convolution. Vanilla dynamic convolution uses a \textit{shared dynamic attention} mechanism to aggregate \textit{unshared static basis vectors} in a \textit{high} dimensional latent space. In contrast, DCD uses an \textit{unshared dynamic channel fusion} mechanism to aggregate \textit{shared static basis vectors} in a \textit{low} dimensional latent space.

\subsection{More General Formulation}

So far, we have focused on the dynamic residual and shown that dynamic channel fusion enables a compact implementation of dynamic convolution. We next discuss the static kernel $\bm{W}_0$. Originally, it is multiplied by a dynamic scalar $\sum_k \pi_k(\bm{x})$, which is canceled in Eq. \ref{eq:dynamic-decom} as attention scores sum to one. Relaxing the constraint $\sum_k \pi_k(\bm{x})=1$ results in the more general form
\begin{align}
\bm{W}(\bm{x})&=\bm{\Lambda(x)}\bm{W}_0 + \bm{P}\bm{\Phi}(\bm{x})\bm{Q}^T, 
\label{eq:dynamic-affine}
\end{align}
where $\bm{\Lambda}(\bm{x})$ is a $C \times C$ diagonal matrix and $\lambda_{i,i}(\bm{x})$ a function of $\bm{x}$. In this way, $\bm{\Lambda}(\bm{x})$ implements channel-wise attention after the static kernel $\bm{W}_0$, generalizing Eq. \ref{eq:dynamic-residual} where $\bm{\Lambda(x)}$ is an identity matrix. Later, we will see that this generalization enables additional performance gains. 

\textbf{Relation to Squeeze-and-Excitation (SE) \citep{Hu_2018_CVPR}:} The dynamic channel-wise attention mechanism implemented by $\bm{\Lambda}(\bm{x})$ is related to but \textit{different} from SE. It is parallel to a convolution and shares the input with the convolution. It can be thought of as either a \textit{dynamic convolution kernel} $\bm{y}=(\bm{\Lambda}(\bm{x})\bm{W}_0)\bm{x}$ or an input-dependent attention mechanism  applied to the \textit{output feature map} of the convolution $\bm{y}=\bm{\Lambda}(\bm{x})(\bm{W}_0\bm{x})$. Thus, its computational complexity is $\min(\mathcal{O}(C^2), \mathcal{O}(HWC))$, where $H$ and $W$ are height and width of the feature map.

In contrast, SE is placed \textit{after} a convolution and uses the output of the convolution as input.
It can only apply channel attention on the \textit{output feature map} of the convolution as $\boldsymbol{y}=\boldsymbol{\Lambda}(\boldsymbol{z})\boldsymbol{z}$, where $\boldsymbol{z}=\boldsymbol{W}_0 \boldsymbol{x}$. Its computational complexity is $\mathcal{O}(HWC)$. Clearly, SE requires more computation than dynamic channel-wise attention $\bm{\Lambda}(\bm{x})$ when the resolution of the feature map ($H \times W$) is high.

\begin{figure}[t]
\begin{minipage}{0.52\linewidth}
\centering
\includegraphics[width=0.99\textwidth]{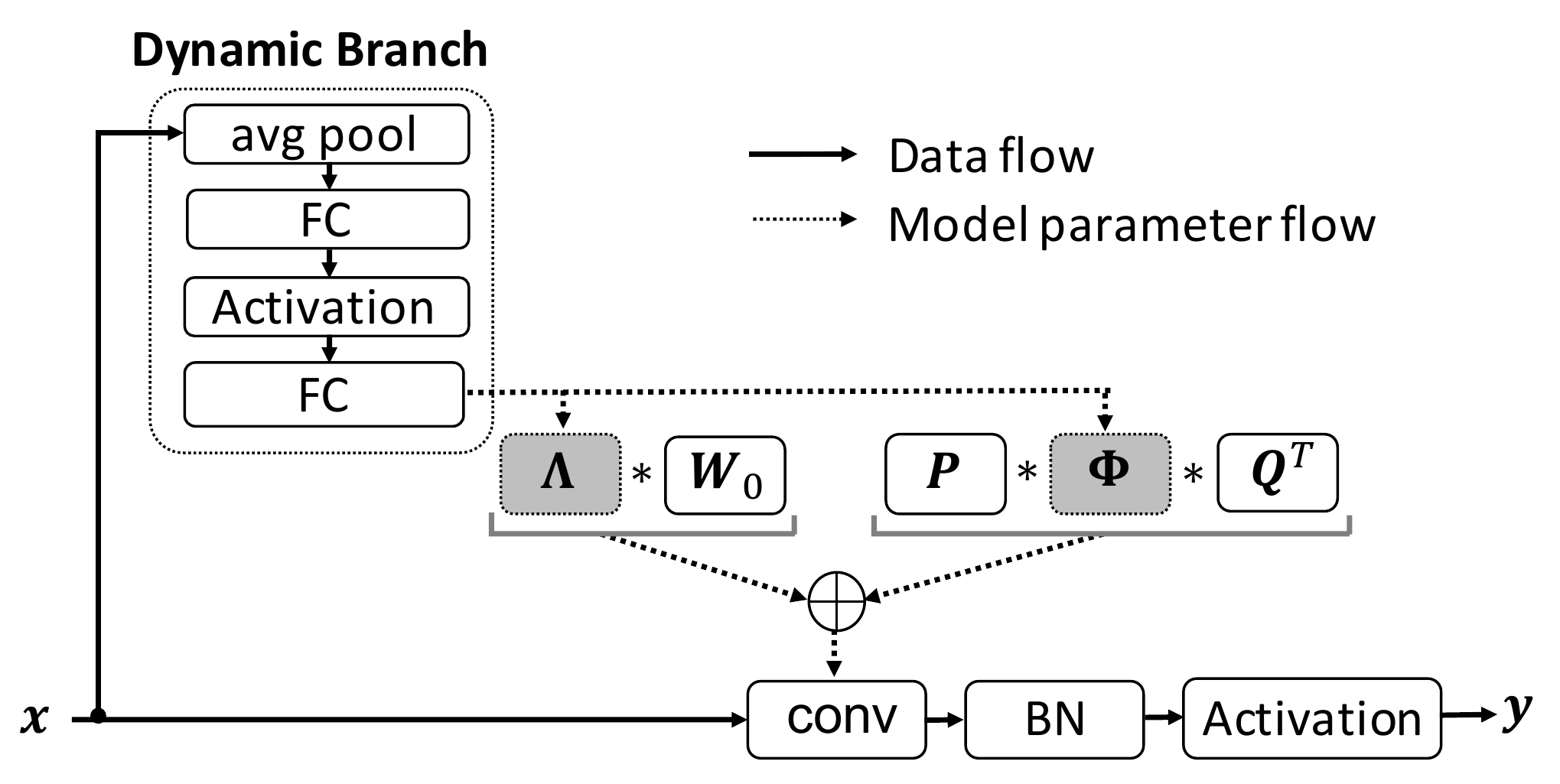}
\caption{\textbf{Dynamic convolution decomposition layer}. The input $\bm{x}$ first goes through a dynamic branch to generate $\bm{\Lambda}(\bm{x})$ and $\bm{\Phi}(\bm{x})$, and then to generate the convolution matrix $\bm{W}(\bm{x})$ 
using Eq. \ref{eq:dynamic-affine}.}
\label{fig:arch}
\end{minipage}
\qquad
\begin{minipage}{0.44\linewidth}
\vspace{2.5em}
\centering
\includegraphics[width=0.99\textwidth]{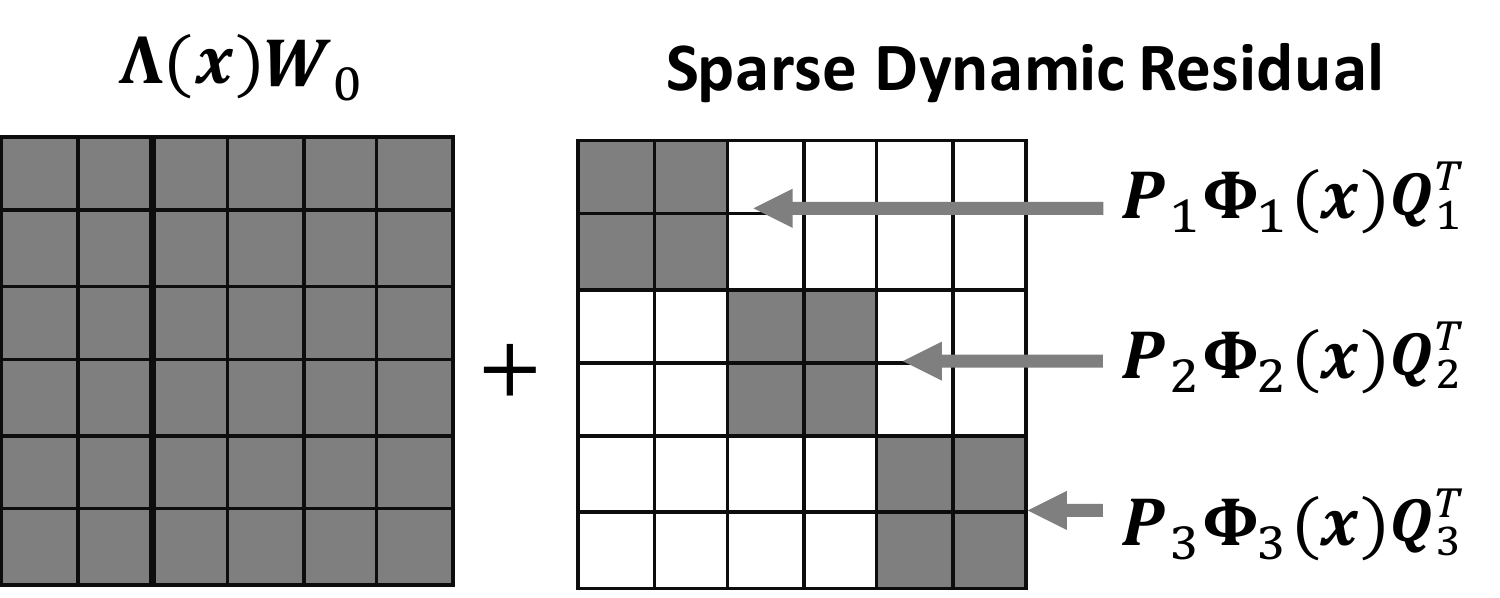}
\caption{\textbf{Sparse dynamic residual}, which is represented as a diagonal block matrix. Each diagonal block is decomposed separately as $\bm{P}_b\bm{\Phi}_b\bm{Q}^T_b$. Note that the static kernel $\bm{W}_0$ is still a full size matrix.}
\label{fig:group}
\end{minipage}
\end{figure}

\subsection{Dynamic Convolution Decomposition Layer}
\textbf{Implementation:} Figure \ref{fig:arch} shows the diagram of a dynamic convolution decomposition (DCD) layer. It uses a light-weight dynamic branch to generate coefficients for both dynamic channel-wise attention $\bm{\Lambda}(\bm{x})$ and dynamic channel fusion $\bm{\Phi}(\bm{x})$. Similar to Squeeze-and-Excitation \citep{Hu_2018_CVPR}, the dynamic branch first 
applies average pooling to the input $\bm{x}$. This is followed by two fully connected (FC) layers with an activation layer between them. The first FC layer reduces the number of channels by $r$ and the second expands them into $C+L^2$ outputs ($C$ for $\bm{\Lambda}$ and $L^2$ for $\bm{\Phi}$).  
Eq. \ref{eq:dynamic-affine} is finally used to generate convolutional weights $\bm{W}(\bm{x})$. Similarly to a static convolution, a DCD layer also includes a batch normalization and an activation (e.g. ReLU) layer.

\textbf{Parameter Complexity:}
DCD has similar FLOPs to the vanilla dynamic convolution. Here, we focus on parameter complexity. 
Static convolution and vanilla dynamic convolution require $C^2$ and $KC^2$ parameters, respectively. 
DCD requires $C^2$, $CL$, and $CL$ parameters for static matrices $\bm{W}_0$, $\bm{P}$ and $\bm{Q}$, respectively. An additional $(2C+L^2)\frac{C}{r}$ parameters are required by the dynamic branch to generate $\bm{\Lambda}(\bm{x})$ and $\bm{\Phi}(\bm{x})$, where $r$ is the reduction rate of the first FC layer. The total complexity is $C^2+2CL+(2C+L^2)\frac{C}{r}$. Since $L$ is constrained as $L^2 < C$, the complexity upper bound is $(1+\frac{3}{r})C^2+2C\sqrt{C}$. When choosing $r=16$, the complexity is about $1\frac{3}{16}C^2$.
This is much less than what is typical for vanilla dynamic convolution ($4C^2$ in \cite{Chen2019DynamicCA}
and $8C^2$ in \cite{Yang2019CondConvCP}).

\section{Extensions of Dynamic Convolution Decomposition}
In this section, we extend the dynamic decomposition of $1 \times 1$ convolution (Eq. \ref{eq:dynamic-affine}) in three ways: (a) sparse dynamic residual where $\bm{P}\bm{\Phi}(\bm{x})\bm{Q}^T$ is a diagonal block matrix, (b) $k \times k$ depthwise convolution, and (c) $k \times k$ convolution. Here, $k$ refers to the kernel size.

\subsection{DCD with Sparse Dynamic Residual}
The dynamic residual $\bm{P}\bm{\Phi}(\bm{x})\bm{Q}^T$ can be further simplified into a block-diagonal matrix of blocks $\bm{P}_b\bm{\Phi}_b(\bm{x})\bm{Q}^T_b, b \in \{1,\ldots, B\}$, leading to
\begin{align}
\bm{W}(\bm{x})&=\bm{\Lambda(x)}\bm{W}_0 + \bigoplus_{b=1}^B \bm{P}_b\bm{\Phi}_b(\bm{x})\bm{Q}^T_b,
\label{eq:dynamic-affine-sparse}
\end{align}
where $\bigoplus_{i=1}^nA_i=diag(A_1,\dots,A_n)$. This form has Eq. \ref{eq:dynamic-affine} as a special case, where $B=1$.
Note that the static kernel $\bm{W}_0$ is still a full matrix and only the dynamic residual is sparse (see Figure \ref{fig:group}). We will show later that keeping as few as $\frac{1}{8}$ of the entries of the dynamic residual non-zero ($B=8$) has a minimal performance degradation, still significantly outperforming a static kernel.

\subsection{DCD of $k \times k$ Depthwise Convolution}
The weights of a $k \times k$ depthwise convolution kernel form a $C \times k^2$ matrix. DCD can be generalized to such matrices by replacing in Eq. \ref{eq:dynamic-affine} the matrix $\bm{Q}$ (which squeezes the number of channels) with a matrix $\bm{R}$ (which squeezes the number of kernel elements) 
\begin{align}
\bm{W}(\bm{x})&=\bm{\Lambda(x)}\bm{W}_0 + \bm{P}\bm{\Phi}(\bm{x})\bm{R}^T,
\label{eq:dynamic-depthwise}
\end{align}
where $\bm{W}(\bm{x})$ and $\bm{W_0}$ are $C \times k^2$ matrices, $\bm{\Lambda(x)}$ is a diagonal $C \times C$ matrix that implements channel-wise attention, $\bm{R}$ is a $k^2 \times L_k$ matrix that reduces the number of kernel elements from $k^2$ to $L_k$, $\bm{\Phi(x)}$ is a $L_k \times L_k$ matrix that performs dynamic fusion along $L_k$ latent kernel elements and $\bm{P}$ is a $C \times L_k$ weight matrix for depthwise convolution over $L_k$ kernel elements. The default value of $L_k$ is $\lfloor k^2/2 \rfloor$. Since depthwise convolution is channel separable, $\bm{\Phi(x)}$ does not fuse channels, fusing instead $L_k$ latent kernel elements.

\subsection{DCD of $k \times k$ Convolution} \label{sec:ext-3x3}

\begin{figure}[t]
\centering
\includegraphics[width=0.95\linewidth]{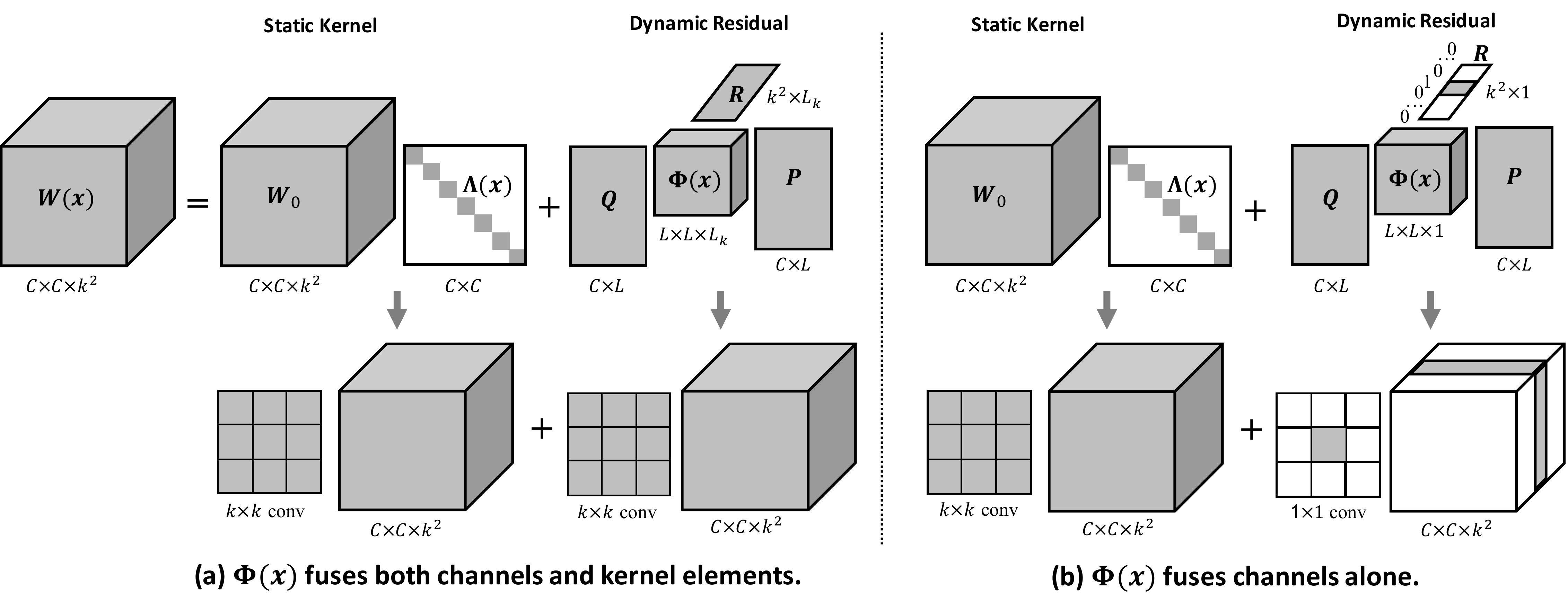}
\caption{The dynamic convolution decomposition for $k \times k$ convolution.}
\label{fig:kxk-tensor}
\vspace{-1.0em}
\end{figure}

\textbf{Joint fusion of channels and kernel elements:} A $k \times k$ convolution kernel forms a $C \times C \times k^2$ tensor. DCD can be generalized to such tensors
by extending  Eq. \ref{eq:dynamic-affine} into a tensor form (see Figure \ref{fig:kxk-tensor})
\begin{align}
\bm{W}(\bm{x})=\bm{W_0} \times_{2}\bm{\Lambda(x)}+\bm{{{\Phi}}}(\bm{x})\times_{1}\bm{Q} \times_{2}\bm{P}\times_{3}\bm{R},
\label{eq:dynamic_res_3d_decom}
\end{align}
where $\times_{n}$ refers to $n$-mode multiplication \citep{Lathauwer2000tensor}, $\bm{W_0}$ is a $C \times C \times k^2$ tensor, $\bm{\Lambda(x)}$ is a diagonal $C \times C$ matrix that implements channel-wise attention,  $\bm{Q}$ is a $C \times L$ matrix that reduces the number of input channels from $C$ to $L$, $\bm{R}$ is a $k^2 \times L_k$ matrix that reduces the number of kernel elements from $k^2$ to $L_k$,
$\bm{\Phi(x)}$ is a $L \times L \times L_k$ tensor that performs joint fusion of $L$ channels over $L_k$ latent kernel elements, and $\bm{P}$ is a $C \times L$ matrix that expands the number of channels from $L$ to $C$. The numbers of latent channels $L$ and latent kernel elements $L_k$ are constrained by $L_k < k^2$ and $L^2L_k \leq C$. Their default values are set empirically to $L_k=\lfloor k^2/2 \rfloor$, $L=\lfloor \frac{C/L_k}{2^{\lfloor log_2\sqrt{C/L_k} \rfloor}} \rfloor$.

\textbf{Channel fusion alone:} We found that the fusion of channels $\bm{{{\Phi}}}(\bm{x}) \times_{1}\bm{Q}$ is more important than the fusion of kernel elements $\bm{{{\Phi}}}(\bm{x})\times_{3}\bm{R}$. Therefore, we reduce $L_k$ to 1 and increase $L$ accordingly. $\bm{R}$ is simplified into a one-hot vector $[0,\dots, 0, 1, 0, \dots, 0]^T$, where the `1' is located at the center (assuming that $k$ is an odd number). As illustrated in Figure \ref{fig:kxk-tensor}-(b), the tensor of dynamic residual $\bm{{{\Phi}}}(\bm{x}) \times_{1}\bm{Q} \times_{2}\bm{P} \times_{3} \bm{R}$ only has one non-zero slice, which is equivalent to a $1 \times 1$ convolution. Therefore, the DCD of a $k \times k$ convolution is essentially adding a $1 \times 1$ dynamic residual to a static $k \times k$ kernel.

\section{Experiments}
In this section, we present the results of DCD on ImageNet classification \citep{deng2009imagenet}. ImageNet has 1,000 classes with 1,281,167  training and $50,000$ validation images. We also report ablation studies on different components of the approach.

All experiments are based on two network architectures: ResNet \citep{he2016deep} and MobileNetV2 \citep{sandler2018mobilenetv2}.
DCD is implemented on all convolutional layers of ResNet and all $1 \times 1$ convolutional layers of MobileNetV2.
The reduction ratio $r$ is set to 16 for ResNet and MobileNetV2 $\times1.0$, and to 8 for smaller models (MobileNetV2 $\times 0.5$ and $\times 0.35$). All models are trained by SGD with momentum 0.9. The batch size is 256 and remaining training parameters are as follows.

{\noindent\bf{ResNet: }}The learning rate starts at $0.1$ and is divided by 10 every 30 epochs. The model is trained with $100$ epochs. Dropout \citep{srivastava2014dropout} $0.1$ is used only for ResNet-50.

{\noindent\bf{MobileNetV2: }}The initial learning rate is $0.05$ and decays to $0$ in $300$ epochs, according to a cosine function. Weight decay of 2e-5 and a dropout rate of $0.1$ are also used. For MobileNetV2 $\times 1.0$, Mixup \citep{zhang2018mixup} and label smoothing are further added to avoid overfitting.

\subsection{Inspecting Different DCD Formulations}
Table \ref{tab:point_conv_decom_res} summarizes the influence of different components (e.g. dynamic channel fusion $\bm{\Phi}(\bm{x})$, dynamic channel-wise attention $\bm{\Lambda}(\bm{x})$) of DCD on MobileNet V2 $\times 0.5$ and ResNet-18 performance.
The table shows that both dynamic components, $\bm{\Lambda}(\bm{x})$ and $\bm{\Phi}(\bm{x})$ of Eq. \ref{eq:dynamic-affine}.
enhance accuracy substantially (+2.8\% and +3.8\% for MobileNetV2 $\times$0.5, +1.1\% and +2.4\% for ResNet-18), when compared to the static baseline. Using dynamic channel fusion only ($\bm{W}_0+\bm{P}\bm{\Phi}\bm{Q}^T$) has slightly more parameters, FLOPs, and accuracy than using dynamic channel-wise attention only ($\bm{\Lambda}\bm{W}_0$). The combination of the two mechanisms provides additional improvement.
\subsection{Ablations}

A number of ablations were performed on MobileNet V2 $\times 0.5$ to analyze DCD performance in terms of two questions.
\begin{enumerate}
    \itemsep0.3em
    \item \textbf{How} does the dimension ($L$) of the latent space affect performance?
    \item \textbf{How} do three DCD variants perform?
\end{enumerate}
The default configuration is the general form of DCD (Eq. \ref{eq:dynamic-affine}) with a full size dynamic residual ($B=1$) for all pointwise convolution layers. The default latent space dimension is $L=\lfloor \frac{C}{2^{\lfloor log_2\sqrt{C} \rfloor}} \rfloor$.

\begin{table}[t]
\caption{
    \textbf{Different formulations} of dynamic convolution decomposition on ImageNet classification.
}
\parbox{.46\linewidth}{
\centering
    \footnotesize
        \begin{tabular}{lcrc}
	    \specialrule{.1em}{.05em}{.05em} 
		Model & Params & MAdds & Top-1\\
		\hline
		$\bm{W}_0$ (static) & 2.0M & 97.0M & 65.4 \\
		\hline
		$\bm{\Lambda}\bm{W}_0$ & 2.4M & 97.4M & 68.2\\
		\hline
		$\bm{W}_0+\bm{P}\bm{\Phi}\bm{Q}^T$ & 2.7M & 104.4M & 69.2 \\
		\hline
		$\bm{\Lambda}\bm{W}_0+\bm{P}\bm{\Phi}\bm{Q}^T$ & 2.9M & 104.6M & \textbf{69.8} \\
		\specialrule{.1em}{.05em}{.05em}
		\\
		\multicolumn{4}{c}{\textbf{(a) MobileNet V2 $\times 0.5$}} \\
		\\
	\end{tabular}
	\label{tab:point_conv_decom_mb}
	\vspace{-0.5em}
}
{
\hfill
\centering
    \footnotesize
        \begin{tabular}{lcrc}
	    \specialrule{.1em}{.05em}{.05em} 
		Model & Params & MAdds & Top-1\\
		\hline
		$\bm{W}_0$ (static) & 11.1M & 1.81G & 70.4 \\
		\hline
		$\bm{\Lambda}\bm{W}_0$ & 11.7M & 1.81G & 71.5\\
		\hline
		$\bm{W}_0+\bm{P}\bm{\Phi}\bm{Q}^T$ & 13.6M & 1.83G & 72.8 \\
		\hline
		$\bm{\Lambda}\bm{W}_0+\bm{P}\bm{\Phi}\bm{Q}^T$ & 14.0M & 1.83G & \textbf{73.1} \\
		\specialrule{.1em}{.05em}{.05em}
		\\
		\multicolumn{4}{c}{\textbf{(b) ResNet-18}} \\
		\\
	\end{tabular}
	\label{tab:point_conv_decom_res}
	\vspace{-2.5em}
}
\end{table}

\begin{wraptable}[13]{r}{0.48\textwidth}
\caption{\textbf{Dimension of the latent space} $L$ evaluated on ImageNet classification (MobileNetV2 $\times$0.5 is used).}
\vspace{-0.3em}
\centering
    \footnotesize
    \begin{tabular}{cccrc}
		\specialrule{.1em}{.05em}{.05em}
		Model & $L$ & Params & MAdds & Top-1 \\
		\hline
		static & - & 2.0M & 97.0M & 65.4\\
		\hline
		\multirow{4}{1.0cm}{\tabincell{c}{DCD}}
		& $\times$0.25 & 2.4M & 99.8M & 68.7\\
		\cline{2-5}
		& $\times$0.50 & 2.5M & 101.3M & 69.0\\
		\cline{2-5}
		&$\times$0.75& 2.6M & 102.9M & 69.6\\
		\cline{2-5}
		& $\times$1.0 & 2.9M & 104.6M & {\bf{69.8}} \\
		\specialrule{.1em}{.05em}{.05em} \\
		\\
	\end{tabular}
	\label{tab:phi_size_ablation}
	\vspace{-2.5em}
\end{wraptable}

\textbf{Latent Space Dimension $L$:}
The dynamic channel fusion matrix $\bm{\Phi}(\bm{x})$ has size $L \times L$. Thus, $L$ controls both the representation and the parameter complexity of DCD. We adjust it by applying different multipliers to the default value of $L$. Table \ref{tab:phi_size_ablation} shows the results of MobileNetV2 $\times$0.5 for four multiplier values ranging from $\times 1.0$ to $\times 0.25$. As $L$ decreases, fewer parameters are required and the performance  degrades slowly. Even with a very low dimensional latent space ($L \times 0.25$), DCD still outperforms the static baseline by 3.3\% top-1 accuracy.

\textbf{Number of Diagonal Blocks $B$ in the Dynamic Residual:}
 Table \ref{tab:group}-(a) shows classification results for four values of $B$. The dynamic residual is a full matrix when $B=1$, while only $\frac{1}{8}$ of its entries are non-zero for $B=8$.
Accuracy degrades slowly as the dynamic residual becomes sparser (increasing $B$).
 The largest performance drop happens when $B$ is changed from 1 to 2, as half of the weight matrix $\bm{W}(\bm{x})$ becomes static. However, performance is still significantly better than that of the static baseline. 
 The fact that even the sparsest $B=8$ outperforms the static baseline by 2.9\% 
 (from 65.4\% to 68.3\%) demonstrates the representation power of the dynamic residual. In all cases, dynamic channel-wise attention $\bm{\Lambda}(\bm{x})$ enables additional performance gains.

\begin{table}[t]
\caption{\textbf{Extensions of dynamic convolution decompostion (DCD)} evaluated on ImageNet classification (MobileNetV2 $\times$0.5 is used).}

\parbox{.48\linewidth}{
\centering
    \footnotesize
	\begin{tabular}{cc@{\hskip 2.8mm}c@{\hskip 2.8mm}r@{\hskip 2.8mm}c}
		\specialrule{.1em}{.05em}{.05em}
		Network & $B$ & Params & MAdds & Top-1 \\
		\hline
		$\bm{W}_0$ (static) & - & 2.0M & 97.0M& 65.4 \\
		\hline
		\multirow{4}{2.0cm}{\tabincell{c}{$\bm{W}_0+\bm{P}\bm{\Phi}\bm{Q}^T$}}
		&1 & 2.7M & 104.4M & {\bf{69.2}} \\
		\cline{2-5}
		&2 & 2.6M & 101.0M & 68.5 \\
		\cline{2-5}
		&4 & 2.5M & 99.1M & 68.4 \\
		\cline{2-5}
		&8 & 2.5M & 98.5M & 68.3 \\
		\hline
		\multirow{4}{2.0cm}{\tabincell{c}{$\bm{\Lambda}\bm{W}_0+\bm{P}\bm{\Phi}\bm{Q}^T$}}
		&1 & 2.9M & 104.6M & {\bf{69.8}} \\
		\cline{2-5}
		&2& 2.8M & 101.3M & 68.9\\
		\cline{2-5}
		&4 & 2.7M & 99.4M & 68.8 \\
		\cline{2-5}
		& 8 & 2.7M & 98.8M & 68.5 \\
		\specialrule{.1em}{.05em}{.05em} \\
		\multicolumn{5}{l}{\shortstack[l]{\textbf{(a) Number of diagonal blocks $B$ in the dynamic} \\ \textbf{residual.}}}
	\end{tabular}
	\label{tab:group}
}
\hfill
\parbox{.49\linewidth}{
\vspace{0.3em}
\centering
    \footnotesize
	\begin{tabular}{c@{\hskip 3.0mm}c@{\hskip 3.0mm}c@{\hskip 3.0mm}c@{\hskip 3.0mm}r@{\hskip 3.0mm}c}
		\specialrule{.1em}{.05em}{.05em}
		DW & PW & CLS & Params & MAdds & Top-1\\
		\hline
		 &  &   & 2.0M & 97.0M & 65.4 \\
		\hline
		\checkmark & & & 2.4M & 97.5M & 68.3 \\
		\hline
		& \checkmark & & 2.9M & 104.6M & 69.8 \\
		\hline
		& & \checkmark & 2.2M & 97.2M & 66.6 \\
		\hline
		\checkmark& & \checkmark & 2.6M & 97.7M & 69.0 \\
		\hline
		\checkmark & \checkmark & & 3.3M & 105.1M & 69.6\\
		\hline
		& \checkmark & \checkmark & 3.1M & 104.8M & {\bf{70.2}} \\
		\hline
		\checkmark & \checkmark & \checkmark & 3.5M & 105.3M & 70.0 \\
		\specialrule{.1em}{.05em}{.05em} \\
		\multicolumn{6}{l}{\shortstack[l]{\textbf{(b) DCD at different layers}. DW, PW, and CLS\\ indicate depthwise convolution, pointwise 
		conv-\\olution and classifier respectively.}} 
	\end{tabular}
	\label{tab:depth_point_conv_decom}
}
\end{table}

\begin{table}[t]
    \caption{Comparing DCD with the vanilla dynamic convolution CondConv \citep{Yang2019CondConvCP} and DY-Conv \citep{Chen2019DynamicCA}. \xmark indicates the dynamic model with the fewest parameters (static model is not included). CondConv contains $K=8$ kernels and DY-Conv contains $K=4$ kernels.}
    \vspace{-1mm}
    \begin{center}
    \footnotesize
    \begin{tabular}{l@{\hskip 3mm}l}
    \begin{tabular}{c@{\hskip 2.5mm}l@{\hskip 2.5mm}r@{\hskip 2.5mm}r@{\hskip 2.5mm}c}
		\specialrule{.1em}{.05em}{.05em}
		Width & Model & Params & MAdds & Top-1 \\
		\hline
		\multirow{4}{*}{\tabincell{c}{$\times 1.0$}}
		&static & 3.5M & 300.0M & 72.0 \\
		&DY-Conv& 11.1M & 312.9M & \textbf{75.2} \\
		&CondConv& 27.5M & 329.0M & 74.6 \\
		&DCD (ours) & \xmark 5.5M & 326.0M & \textbf{75.2} \\
		\hline
		\multirow{3}{*}{\tabincell{c}{$\times 0.5$}}
		&static& 2.0M & 97.0M & 65.4 \\
		&DY-Conv& 4.0M & 101.4M & 69.9 \\
		&CondConv& 15.5M & 113.0M & 68.4 \\
		&DCD (ours)& \xmark 3.1M & 104.8M & \textbf{70.2} \\
		\hline
		\multirow{3}{*}{\tabincell{c}{$\times 0.35$}}
		&static& 1.7M & 59.2M & 60.3 \\
		&DY-Conv& 2.8M & 62.0M & 65.9 \\
		&DCD (ours)& \xmark 2.3M & 63.1M & \textbf{66.6} \\
		\specialrule{.1em}{.05em}{.05em} \\
		\multicolumn{5}{c}{(a) \textbf{MobileNetV2}.}
    \end{tabular}
&
    \begin{tabular}{c@{\hskip 2.5mm}l@{\hskip 2.5mm}r@{\hskip 2.5mm}r@{\hskip 2.5mm}c}
    \\
    \\
    \\
		\specialrule{.1em}{.05em}{.05em}
		Depth & Model & Params & MAdds & Top-1 \\
		\hline
		\multirow{2}{*}{\tabincell{c}{ResNet-50}}
		&static &23.5M & 3.8G&76.2 \\
		&DCD (ours)&30.7M & 3.9G& \textbf{77.9} \\
		\hline
		\multirow{3}{*}{\tabincell{c}{ResNet-18}}
		&static &11.1M &1.81G &70.4 \\
		&DY-Conv &42.7M &1.85G &72.7 \\
		&DCD (ours)&\xmark 14.0M &1.83G & \textbf{73.1} \\
		\hline
        \multirow{3}{*}{\tabincell{c}{ResNet-10}}
		&static &5.2M &0.89G &63.5 \\
		&DY-Conv &18.6M &0.91G &67.7 \\
		&DCD (ours)& \xmark 6.5M &0.90G & \textbf{68.8} \\
		\specialrule{.1em}{.05em}{.05em} \\
		\multicolumn{5}{c}{(b) \textbf{ResNet}.}
    \end{tabular}
    \end{tabular}
    \end{center}
    \label{tab:main_res}
    \vspace{-5mm}
\end{table}

\textbf{DCD at Different Layers:}
Table \ref{tab:depth_point_conv_decom}-(b) shows the results of implementing DCD for three different types of layers (a) DW: depthwise convolution (Eq. \ref{eq:dynamic-depthwise}), (b) PW: pointwise convolution (Eq. \ref{eq:dynamic-affine}), and (c) CLS: fully connected classifier, which is a special case of pointwise convolution (the input resolution is $1 \times 1$). 
Using DCD in any type of layer improves on the performance of the static baseline (+2.9\% for depthwise convolution, +4.4\% for pointwise convolution, and +1.2\% for classifier). Combining DCD for both pointwise convolution and classifier achieves the best performance (+4.8\%). We notice a performance drop (from 70.2\% to 70.0\%) when using DCD in all three types of layers. We believe this is due to overfitting, as it has higher training accuracy. 

\textbf{Extension to $3 \times 3$ Convolution:}
We use ResNet-18, which stacks 16 layers of $3 \times 3$ convolution, to study the $3 \times 3$ extension of DCD (see Section \ref{sec:ext-3x3}). Compared to the static baseline (70.4\% top-1 accuracy), DCD with \textit{joint fusion of channels and kernel elements} (Eq. \ref{eq:dynamic_res_3d_decom}) improves top-1 accuracy (71.3\%) by 0.9\%. The top-1 accuracy is further improved by 1.8\% (73.1\%), when using DCD with \textit{channel fusion alone}, 
which transforms the dynamic residual as a $1 \times 1$ convolution matrix (see Figure \ref{fig:kxk-tensor}-(b)). This demonstrates that dynamic fusion is more effective across channels than across kernel elements.

\textbf{Summary:} Based on the ablations above, DCD should be implemented with both dynamic channel fusion $\bm{\Phi}$ and dynamic channel-wise attention $\bm{\Lambda}$, the default latent space dimension $L$, and a full size residual $B=1$. DCD is recommended for pointwise convolution and classifier layers in MobileNetV2. For $3 \times 3$ convolutions in ResNet, DCD should be implemented with channel fusion alone. 
The model can be made more compact, for a slight performance drop, by (a) removing dynamic channel-wise attention $\bm{\Lambda}$, (b) reducing the latent space dimension $L$, (c) using a sparser dynamic residual (increasing $B$), and (d) implementing DCD in depthwise convolution alone.  

\subsection{Main Results}

DCD was compared to the vanilla dynamic convolution \citep{Yang2019CondConvCP, Chen2019DynamicCA} for MobileNetV2 and ResNet, using the settings recommended above, with the results of
\begin{wrapfigure}[17]{r}{0.38\textwidth}
\centering
\includegraphics[width=.98\linewidth]{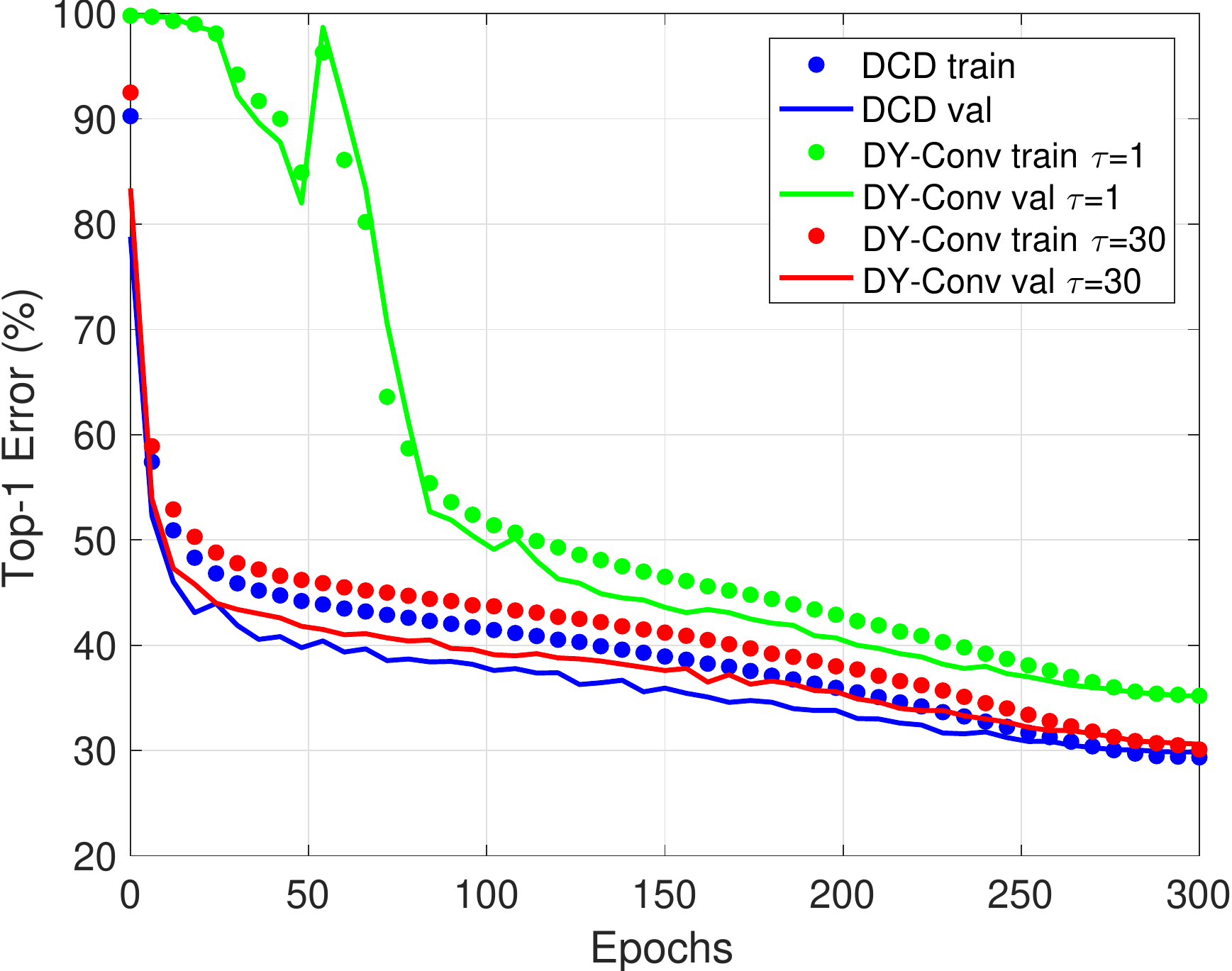}
\caption{The comparison of training and validation error between DCD and DY-Conv on MobileNetV2 $\times0.5$. $\tau$ is the temperature in softmax. Best viewed in color.}
\label{Fig:Data2}
\end{wrapfigure}
 Table \ref{tab:main_res}\footnote{The baseline results are from the original papers. Our implementation, under the setup used for DCD, has either similar or slightly lower results, e.g. for MobileNetV2$\times$1.0 the original paper reports 72.0\%, while our implementation achieves 71.8\%.}.
 DCD significantly reduces the number of parameters while improving the performance of both network architectures. For MobileNetV2 $\times 1.0$, DCD only requires 50\% of the parameters of \citep{Chen2019DynamicCA} and 25\% of the parameters of \citep{Yang2019CondConvCP}. For ResNet-18, it only requires 33\% of the parameters of \citep{Chen2019DynamicCA}, while achieving a 0.4\% gain in top-1 accuracy. Although DCD requires slightly more MAdds than \citep{Chen2019DynamicCA}, the increment is negligible. These results demonstate that DCD is more compact and effective.

Figure \ref{Fig:Data2} compares DCD to DY-Conv \citep{Chen2019DynamicCA} in terms of training convergence. DY-Conv uses a large temperature in its softmax to alleviate the joint optimization difficulty and make training more efficient. Without any additional parameter tuning, DCD converges even faster than DY-Conv with a large temperature and achieves higher accuracy. 

\subsection{Analysis of Dynamic Channel Fusion}
\begin{wrapfigure}[12]{r}{0.38\textwidth}
\vspace{-2.5em}
\centering
\includegraphics[width=.95\linewidth]{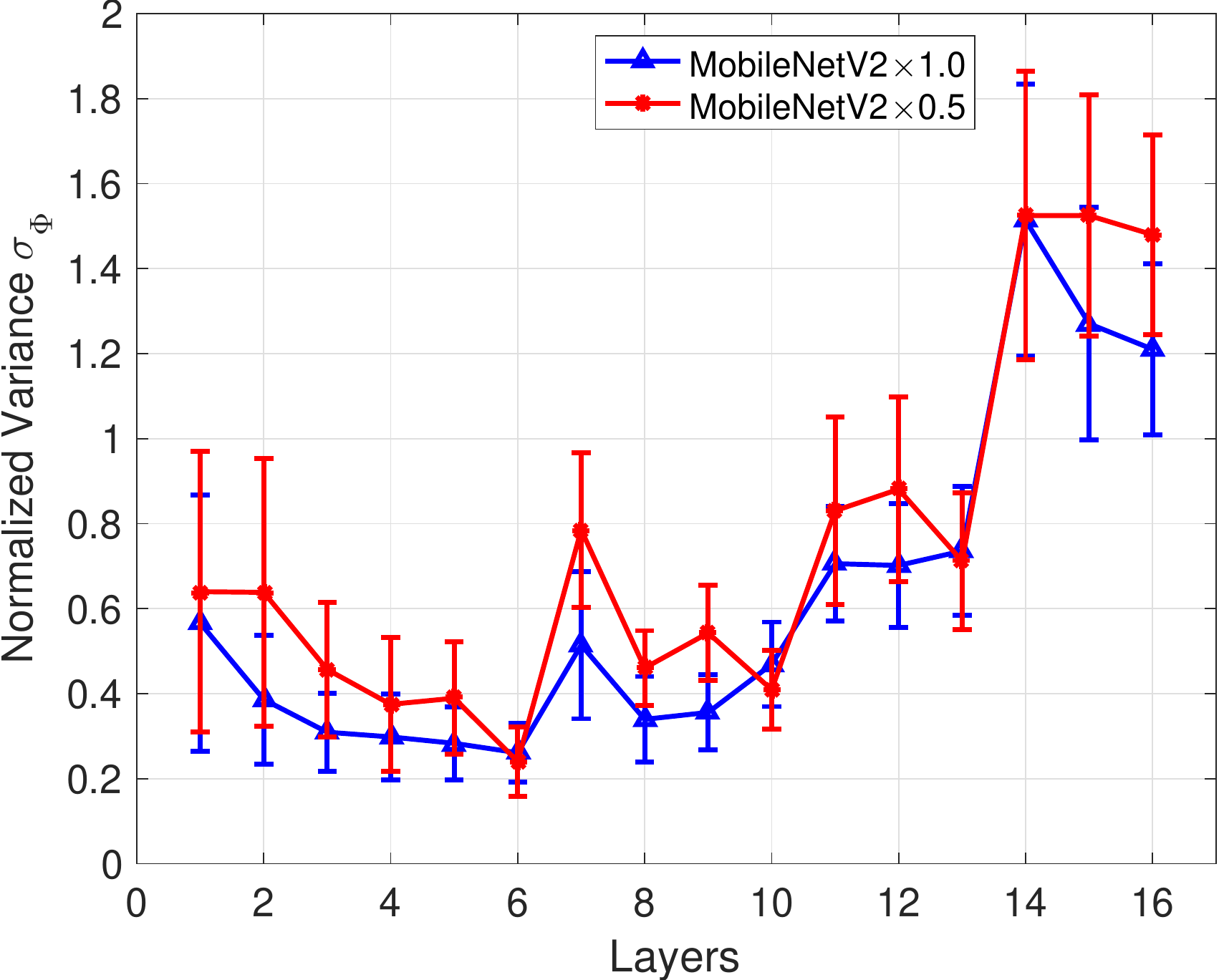}
\caption{Normalized variance of dynamic coefficients $\sigma_{\Phi}$ across layers in MobileNetV2 $\times0.5$ and $\times1.0$.}
\label{fig:var-ratio}
\end{wrapfigure}
To validate the \textit{dynamic} property, $\bm{\Phi}(\bm{x})$ should have different values over different images. We measure this by averaging the variance of each entry $\sigma_\Phi=\sum_{i,j}\sigma_{i,j}/L^2$, where $\sigma_{i,j}$ is the variance of $\phi_{i,j}(\bm{x})$, over all validation images. To compare $\sigma_\Phi$ across layers, we normalize it by the variance of the corresponding input feature map. 
Figure \ref{fig:var-ratio} shows the normalized variance $\sigma_\Phi$ across layers in MobileNetV2. Clearly, the dynamic coefficients vary more in the higher layers. We believe this is because the higher layers encode more context information, providing more clues to adapt convolution weights.

\subsection{Inference Time}
We use a single-threaded core AMD EPYC CPU 7551P (2.0 GHz) to measure running time (in milliseconds) on MobileNetV2 $\times0.5$ and $\times1.0$. Running time is calculated by averaging the inference time of 5,000 images with batch size 1. Both static baseline and DCD are implemented in PyTorch. Compared with the static baseline, DCD consumes about 8\% more MAdds (97.0M vs 104.8M) and 14\% more running time (91ms vs 104ms) for MobileNetV2 $\times0.5$. For MobileNetV2 $\times1.0$, DCD consumes 9\% more MAdds (300.0M vs 326.0M) and 12\% more running time (146ms vs 163ms). The overhead is higher in running time than MAdds. We believe this is because the optimizations of global average pooling and fully connected layers are not as efficient as convolution. This small penalty in inference time is justified by the DCD gains of 4.8\% and 3.2\% top-1 accuracy over MobileNetV2 $\times0.5$ and $\times1.0$ respectively.

\section{Conclusion}
In this paper, we have revisited dynamic convolution via matrix decomposition and demonstrated the limitations of dynamic attention over channel groups: it multiplies the number of parameters by $K$ and increases the difficulty of joint optimization. We proposed a dynamic convolution decomposition to address these issues. This applies dynamic channel fusion to significantly reduce the dimensionality of the latent space, resulting in a more compact model that is easier to learn with often improved accuracy. We hope that our work provides a deeper understanding of the gains recently observed for dynamic convolution.  

\bibliography{iclr2021_conference}

\begin{thebibliography}{35}
\providecommand{\natexlab}[1]{#1}
\providecommand{\url}[1]{\texttt{#1}}
\expandafter\ifx\csname urlstyle\endcsname\relax
  \providecommand{\doi}[1]{doi: #1}\else
  \providecommand{\doi}{doi: \begingroup \urlstyle{rm}\Url}\fi

\bibitem[Cai et~al.(2019)Cai, Gan, and Han]{Cai2019OnceFA}
Han Cai, Chuang Gan, and Song Han.
\newblock Once for all: Train one network and specialize it for efficient
  deployment.
\newblock \emph{ArXiv}, abs/1908.09791, 2019.

\bibitem[Chen et~al.(2020{\natexlab{a}})Chen, Wang, Xu, Shi, Xu, Tian, and
  Xu]{Chen_2020_CVPR_addernet}
Hanting Chen, Yunhe Wang, Chunjing Xu, Boxin Shi, Chao Xu, Qi~Tian, and Chang
  Xu.
\newblock Addernet: Do we really need multiplications in deep learning?
\newblock In \emph{Proceedings of the IEEE/CVF Conference on Computer Vision
  and Pattern Recognition (CVPR)}, June 2020{\natexlab{a}}.

\bibitem[Chen et~al.(2020{\natexlab{b}})Chen, Wang, Guo, Zhang, and
  Sun]{Chen2020DynamicRC}
Jyun-Ruei Chen, Xijun Wang, Zichao Guo, X.~Zhang, and J.~Sun.
\newblock Dynamic region-aware convolution.
\newblock \emph{ArXiv}, abs/2003.12243, 2020{\natexlab{b}}.

\bibitem[Chen et~al.(2020{\natexlab{c}})Chen, Dai, Liu, Chen, Yuan, and
  Liu]{Chen2019DynamicCA}
Yinpeng Chen, Xiyang Dai, Mengchen Liu, Dongdong Chen, Lu~Yuan, and Zicheng
  Liu.
\newblock Dynamic convolution: Attention over convolution kernels.
\newblock In \emph{IEEE Conference on Computer Vision and Pattern Recognition
  (CVPR)}, 2020{\natexlab{c}}.

\bibitem[Chen et~al.(2020{\natexlab{d}})Chen, Dai, Liu, Chen, Yuan, and
  Liu]{Chen2020DynamicReLU}
Yinpeng Chen, Xiyang Dai, Mengchen Liu, Dongdong Chen, Lu~Yuan, and Zicheng
  Liu.
\newblock Dynamic relu.
\newblock \emph{arXiv preprint arXiv:2003.10027}, abs/2003.10027,
  2020{\natexlab{d}}.

\bibitem[Deng et~al.(2009)Deng, Dong, Socher, Li, Li, and
  Fei-Fei]{deng2009imagenet}
Jia Deng, Wei Dong, Richard Socher, Li-Jia Li, Kai Li, and Li~Fei-Fei.
\newblock Imagenet: A large-scale hierarchical image database.
\newblock In \emph{2009 IEEE conference on computer vision and pattern
  recognition}, pp.\  248--255. Ieee, 2009.

\bibitem[Denton et~al.(2014)Denton, Zaremba, Bruna, LeCun, and
  Fergus]{denton2014exploiting}
Emily~L Denton, Wojciech Zaremba, Joan Bruna, Yann LeCun, and Rob Fergus.
\newblock Exploiting linear structure within convolutional networks for
  efficient evaluation.
\newblock In \emph{Advances in neural information processing systems}, pp.\
  1269--1277, 2014.

\bibitem[Ha et~al.(2017)Ha, Dai, and Le]{Ha2017HyperNetworks}
David Ha, Andrew~M. Dai, and Quoc~V. Le.
\newblock Hypernetworks.
\newblock \emph{ICLR}, 2017.

\bibitem[Han et~al.(2020)Han, Wang, Tian, Guo, Xu, and
  Xu]{Han_2020_CVPR_ghostnet}
Kai Han, Yunhe Wang, Qi~Tian, Jianyuan Guo, Chunjing Xu, and Chang Xu.
\newblock Ghostnet: More features from cheap operations.
\newblock In \emph{IEEE/CVF Conference on Computer Vision and Pattern
  Recognition (CVPR)}, June 2020.

\bibitem[He et~al.(2016)He, Zhang, Ren, and Sun]{he2016deep}
Kaiming He, Xiangyu Zhang, Shaoqing Ren, and Jian Sun.
\newblock Deep residual learning for image recognition.
\newblock In \emph{Proceedings of the IEEE conference on computer vision and
  pattern recognition}, pp.\  770--778, 2016.

\bibitem[Howard et~al.(2019)Howard, Sandler, Chu, Chen, Chen, Tan, Wang, Zhu,
  Pang, Vasudevan, Le, and Adam]{howard2019mbnetv3}
Andrew Howard, Mark Sandler, Grace Chu, Liang{-}Chieh Chen, Bo~Chen, Mingxing
  Tan, Weijun Wang, Yukun Zhu, Ruoming Pang, Vijay Vasudevan, Quoc~V. Le, and
  Hartwig Adam.
\newblock Searching for mobilenetv3.
\newblock \emph{CoRR}, abs/1905.02244, 2019.
\newblock URL \url{http://arxiv.org/abs/1905.02244}.

\bibitem[Howard et~al.(2017)Howard, Zhu, Chen, Kalenichenko, Wang, Weyand,
  Andreetto, and Adam]{howard2017mobilenets}
Andrew~G Howard, Menglong Zhu, Bo~Chen, Dmitry Kalenichenko, Weijun Wang,
  Tobias Weyand, Marco Andreetto, and Hartwig Adam.
\newblock Mobilenets: Efficient convolutional neural networks for mobile vision
  applications.
\newblock \emph{arXiv preprint arXiv:1704.04861}, 2017.

\bibitem[Hu et~al.(2018)Hu, Shen, and Sun]{Hu_2018_CVPR}
Jie Hu, Li~Shen, and Gang Sun.
\newblock Squeeze-and-excitation networks.
\newblock In \emph{The IEEE Conference on Computer Vision and Pattern
  Recognition (CVPR)}, June 2018.

\bibitem[Jarrett et~al.(2009)Jarrett, Kavukcuoglu, Ranzato, and
  LeCun]{JarrettKRL09Relu}
Kevin Jarrett, Koray Kavukcuoglu, Marc'Aurelio Ranzato, and Yann LeCun.
\newblock What is the best multi-stage architecture for object recognition?
\newblock In \emph{The IEEE International Conference on Computer Vision
  (ICCV)}, 2009.

\bibitem[Kim et~al.(2015)Kim, Park, Yoo, Choi, Yang, and
  Shin]{kim2015compression}
Yong-Deok Kim, Eunhyeok Park, Sungjoo Yoo, Taelim Choi, Lu~Yang, and Dongjun
  Shin.
\newblock Compression of deep convolutional neural networks for fast and low
  power mobile applications.
\newblock \emph{arXiv preprint arXiv:1511.06530}, 2015.

\bibitem[Kossaifi et~al.(2020)Kossaifi, Toisoul, Bulat, Panagakis, Hospedales,
  and Pantic]{kossaifi2020factorized}
Jean Kossaifi, Antoine Toisoul, Adrian Bulat, Yannis Panagakis, Timothy~M
  Hospedales, and Maja Pantic.
\newblock Factorized higher-order cnns with an application to spatio-temporal
  emotion estimation.
\newblock In \emph{Proceedings of the IEEE/CVF Conference on Computer Vision
  and Pattern Recognition}, pp.\  6060--6069, 2020.

\bibitem[Lathauwer et~al.(2000)Lathauwer, Moor, and
  Vandewalle]{Lathauwer2000tensor}
L.~D. Lathauwer, B.~D. Moor, and J.~Vandewalle.
\newblock A multilinear singular value decomposition.
\newblock In \emph{SIAM J. Matrix Anal. Appl}, 2000.

\bibitem[Lebedev et~al.(2014)Lebedev, Ganin, Rakhuba, Oseledets, and
  Lempitsky]{lebedev2014speeding}
Vadim Lebedev, Yaroslav Ganin, Maksim Rakhuba, Ivan Oseledets, and Victor
  Lempitsky.
\newblock Speeding-up convolutional neural networks using fine-tuned
  cp-decomposition.
\newblock \emph{arXiv preprint arXiv:1412.6553}, 2014.

\bibitem[Li et~al.(2019)Li, Wang, Hu, and Yang]{Li_2019_CVPR_SKNet}
Xiang Li, Wenhai Wang, Xiaolin Hu, and Jian Yang.
\newblock Selective kernel networks.
\newblock In \emph{Proceedings of the IEEE/CVF Conference on Computer Vision
  and Pattern Recognition (CVPR)}, June 2019.

\bibitem[Ma et~al.(2018)Ma, Zhang, Zheng, and Sun]{ma_2018_ECCV}
Ningning Ma, Xiangyu Zhang, Hai-Tao Zheng, and Jian Sun.
\newblock Shufflenet v2: Practical guidelines for efficient cnn architecture
  design.
\newblock In \emph{The European Conference on Computer Vision (ECCV)},
  September 2018.

\bibitem[Ma et~al.(2020)Ma, Zhang, Huang, and Sun]{Ma_2020_eccv_WeightNetRT}
Ningning Ma, X.~Zhang, J.~Huang, and J.~Sun.
\newblock Weightnet: Revisiting the design space of weight networks.
\newblock volume abs/2007.11823, 2020.

\bibitem[Nair \& Hinton(2010)Nair and Hinton]{NairH10Relu}
Vinod Nair and Geoffrey~E. Hinton.
\newblock Rectified linear units improve restricted boltzmann machines.
\newblock In \emph{ICML}, 2010.

\bibitem[Phan et~al.(2020)Phan, Sobolev, Sozykin, Ermilov, Gusak, Tichavsky,
  Glukhov, Oseledets, and Cichocki]{phan2020stable}
Anh-Huy Phan, Konstantin Sobolev, Konstantin Sozykin, Dmitry Ermilov, Julia
  Gusak, Petr Tichavsky, Valeriy Glukhov, Ivan Oseledets, and Andrzej Cichocki.
\newblock Stable low-rank tensor decomposition for compression of convolutional
  neural network.
\newblock \emph{arXiv preprint arXiv:2008.05441}, 2020.

\bibitem[Sandler et~al.(2018)Sandler, Howard, Zhu, Zhmoginov, and
  Chen]{sandler2018mobilenetv2}
Mark Sandler, Andrew Howard, Menglong Zhu, Andrey Zhmoginov, and Liang-Chieh
  Chen.
\newblock Mobilenetv2: Inverted residuals and linear bottlenecks.
\newblock In \emph{Proceedings of the IEEE Conference on Computer Vision and
  Pattern Recognition}, pp.\  4510--4520, 2018.

\bibitem[Srivastava et~al.(2014)Srivastava, Hinton, Krizhevsky, Sutskever, and
  Salakhutdinov]{srivastava2014dropout}
Nitish Srivastava, Geoffrey Hinton, Alex Krizhevsky, Ilya Sutskever, and Ruslan
  Salakhutdinov.
\newblock Dropout: A simple way to prevent neural networks from overfitting.
\newblock \emph{Journal of Machine Learning Research}, 15\penalty0
  (56):\penalty0 1929--1958, 2014.
\newblock URL \url{http://jmlr.org/papers/v15/srivastava14a.html}.

\bibitem[Su et~al.(2020)Su, Fang, xiong Kang, Hu, Pietik{\"a}inen, and
  Liu]{Su_2020_eccv_DynamicGC}
Zhuo Su, Linpu Fang, Wen xiong Kang, D.~Hu, M.~Pietik{\"a}inen, and Li~Liu.
\newblock Dynamic group convolution for accelerating convolutional neural
  networks.
\newblock In \emph{ECCV}, August 2020.

\bibitem[Tan \& Le(2019{\natexlab{a}})Tan and Le]{tan-ICML19-efficientnet}
Mingxing Tan and Quoc Le.
\newblock Efficientnet: Rethinking model scaling for convolutional neural
  networks.
\newblock In \emph{ICML}, pp.\  6105--6114, Long Beach, California, USA, 09--15
  Jun 2019{\natexlab{a}}.

\bibitem[Tan \& Le(2019{\natexlab{b}})Tan and Le]{Tan-bmvc2019-mixconv}
Mingxing Tan and Quoc~V. Le.
\newblock Mixconv: Mixed depthwise convolutional kernels.
\newblock In \emph{30th British Machine Vision Conference 2019},
  2019{\natexlab{b}}.

\bibitem[Tan et~al.(2020)Tan, Pang, and Le]{Tan_2020_CVPR}
Mingxing Tan, Ruoming Pang, and Quoc~V. Le.
\newblock Efficientdet: Scalable and efficient object detection.
\newblock In \emph{Proceedings of the IEEE/CVF Conference on Computer Vision
  and Pattern Recognition (CVPR)}, June 2020.

\bibitem[Tian et~al.(2020)Tian, Shen, and Chen]{Tian_2020_eccv_ConditionalCF}
Zhi Tian, Chunhua Shen, and Hao Chen.
\newblock Conditional convolutions for instance segmentation.
\newblock In \emph{ECCV}, August 2020.

\bibitem[Yang et~al.(2019)Yang, Bender, Le, and Ngiam]{Yang2019CondConvCP}
Brandon Yang, Gabriel Bender, Quoc~V. Le, and Jiquan Ngiam.
\newblock Condconv: Conditionally parameterized convolutions for efficient
  inference.
\newblock In \emph{NeurIPS}, 2019.

\bibitem[Yu et~al.(2019)Yu, Yang, Xu, Yang, and Huang]{yu2018slimmable}
Jiahui Yu, Linjie Yang, Ning Xu, Jianchao Yang, and Thomas Huang.
\newblock Slimmable neural networks.
\newblock In \emph{International Conference on Learning Representations}, 2019.
\newblock URL \url{https://openreview.net/forum?id=H1gMCsAqY7}.

\bibitem[Zhang et~al.(2018{\natexlab{a}})Zhang, Cisse, Dauphin, and
  Lopez-Paz]{zhang2018mixup}
Hongyi Zhang, Moustapha Cisse, Yann~N. Dauphin, and David Lopez-Paz.
\newblock mixup: Beyond empirical risk minimization.
\newblock In \emph{International Conference on Learning Representations},
  2018{\natexlab{a}}.
\newblock URL \url{https://openreview.net/forum?id=r1Ddp1-Rb}.

\bibitem[Zhang et~al.(2018{\natexlab{b}})Zhang, Zhou, Lin, and
  Sun]{Zhang_2018_CVPR}
Xiangyu Zhang, Xinyu Zhou, Mengxiao Lin, and Jian Sun.
\newblock Shufflenet: An extremely efficient convolutional neural network for
  mobile devices.
\newblock In \emph{The IEEE Conference on Computer Vision and Pattern
  Recognition (CVPR)}, June 2018{\natexlab{b}}.

\bibitem[Zhou et~al.(2020)Zhou, Hou, Chen, Feng, and
  Yan]{Daquan_2020_ECCV_RethinkingBS}
Daquan Zhou, Qi-Bin Hou, Y.~Chen, Jiashi Feng, and S.~Yan.
\newblock Rethinking bottleneck structure for efficient mobile network design.
\newblock In \emph{ECCV}, August 2020.

\end{thebibliography}
\bibliographystyle{iclr2021_conference}
\end{document}